\documentclass[sigconf]{acmart}

\makeatletter                   
\def\mdseries@tt{m}             
\makeatother                    

\usepackage{underscore}           
\usepackage{amsmath}
\usepackage{latexsym}
\usepackage{subcaption}
\usepackage{algorithm}
\usepackage{algpseudocode}
\usepackage{multirow}
\usepackage{multicol}
\usepackage{booktabs}
\usepackage[newfloat=true,frozencache=true]{minted} 
\usepackage{color}
\usepackage{listings}
\usepackage{graphicx}
\usepackage{dblfloatfix}
\usepackage{hyperref}           
\hypersetup{
    colorlinks=true,
    linkcolor=blue,
    filecolor=red,      
    urlcolor=magenta,
    breaklinks=true,            
}
\usepackage{breakurl}           

\global\long\def\argmax{\operatornamewithlimits{arg\, max}}

\setcopyright{none} 
\acmConference[Anonymous Submission to AISEC Workshop 2018]{AISEC Workshop at ACM Conference on Computer and Communications Security 2018}{Due 2 July 2018}{Toronto, Canada}
\acmYear{2018}

\newif\ifsubmit
\submitfalse
\ifsubmit
\newcommand{\sknote}[1]{}

\else
\newcommand{\sknote}[1]{\textcolor{blue}{\textbf{Sidd: #1}}}

\fi

\global\long\def\argmax{\operatornamewithlimits{arg\, max}}

\algnewcommand{\IfThen}[2]{
  \State \algorithmicif\ #1\ \algorithmicthen\ #2}

\makeatletter
\renewcommand\fs@ruled{%
  \def\@fs@cfont{\rmfamily}%
  \let\@fs@capt\floatc@plain%
  \def\@fs@pre{\hrule height.8pt depth0pt \kern2pt}
  \def\@fs@post{}
  \def\@fs@mid{\kern2pt\hrule\kern2pt}
  \let\@fs@iftopcapt\iffalse}
\makeatother

\begin{document}
\sloppy
\title{Adaptive Grey-Box Fuzz-Testing with Thompson Sampling}

\author{Siddharth Karamcheti}
\authornote{Work completed while an intern at Bloomberg}
\affiliation{%
	\institution{Bloomberg}
    \department{CTO Data Science}
    \city{New York}
    \state{NY}
    \country{USA}
}
\email{sidd.karamcheti@gmail.com}
\email{}

\author{Gideon Mann}
\affiliation{%
	\institution{Bloomberg}
    \department{CTO Data Science}
    \city{New York}
    \state{NY}
    \country{USA}
}
\email{gmann16@bloomberg.net}

\author{David Rosenberg}
\affiliation{%
	\institution{Bloomberg}
    \department{CTO Data Science}
    \city{New York}
    \state{NY}
    \country{USA}
}
\email{drosenberg44@bloomberg.net}

\begin{abstract}

Fuzz testing, or ``fuzzing,'' refers to a widely deployed class of techniques for testing programs by generating a set of inputs for the express purpose of finding bugs and identifying security flaws. Grey-box fuzzing, the most popular fuzzing strategy, combines light program instrumentation with a data driven process to generate new program inputs. In this work, we present a machine learning approach that builds on AFL, the preeminent grey-box fuzzer, by adaptively learning a probability distribution over its mutation operators on a program-specific basis. These operators, which are selected uniformly at random in AFL and mutational fuzzers in general, dictate how new inputs are generated, a core part of the fuzzer's efficacy. Our main contributions are two-fold: First, we show that a sampling distribution over mutation operators estimated from training programs can significantly improve performance of AFL. Second, we introduce a Thompson Sampling, bandit-based optimization approach that fine-tunes the mutator distribution adaptively, during the course of fuzzing an \textit{individual} program. A set of experiments across complex programs demonstrates that tuning the mutational operator distribution generates sets of inputs that yield significantly higher code coverage and finds more crashes faster and more reliably than both baseline versions of AFL as well as other AFL-based learning approaches.

\end{abstract}



\copyrightyear{2018} 
\acmYear{2018} 
\setcopyright{acmlicensed}
\acmConference[AISec '18]{11th ACM Workshop on Artificial Intelligence and Security}{October 19, 2018}{Toronto, ON, Canada}
\acmBooktitle{11th ACM Workshop on Artificial Intelligence and Security (AISec '18), October 19, 2018, Toronto, ON, Canada}
\acmPrice{15.00}
\acmDOI{10.1145/3270101.3270108}
\acmISBN{978-1-4503-6004-3/18/10}


\maketitle

\section{Introduction}

\begin{algorithm}[t]
  \begin{algorithmic}[1]
      \State \emph{//} \textbf{Core Algorithm for American Fuzzy Lop (AFL)}
      \State \emph{// time: Fixed time window to fuzz (e.g. 24 hours)}
      \State \emph{// queue: Queue of inputs that exercise new code paths.}
      \While{\emph{time} has not elapsed}
      	\State $parent, energy \gets \tt{pick\_input(} \emph{queue} \tt{)}$
      \For{$i \in \bf{range(\emph{energy})}$} 
      	
        \State $child \gets parent$
        \For{$j \in 1$ to $\tt{sample\_num\_mutations()}$}
        	\State $mutation \gets \tt{sample\_mutation()}$
            \State $site \gets \tt{sample\_mutation\_site()}$
            \State $child \gets \tt{apply\_mutation(} \emph{mutation, child, site} \tt{)}$
        \EndFor
        \State $path \gets \tt{execute\_path(} \emph{child, code} \tt{)}$
        \IfThen{($path$ is new)}{$queue \gets child$}
      \EndFor
      \EndWhile
  \end{algorithmic}
  \caption{\textbf{AFL Grey-Box Mutational Fuzzing Algorithm \medskip \\ While $\texttt{pick\_input}$ picks parent seeds and energy intelligently (based on discovery time, execution time, etc.), the other fuzzing parameters (Functions $\texttt{sample\_mutation}$, $\texttt{sample\_num\_mutations}$, and $\texttt{sample\_mutation\_site}$) are random, leading to inefficiencies.}}
  \label{alg:greybox}
\end{algorithm}

\textbf{Program fuzz-testing}, or \textbf{``fuzzing.''} \cite{Godefroid2012SAGEWF, Miller1990AnES,Cha2015ProgramAdaptiveMF,Rebert2014OptimizingSS,Ganesh2009TaintbasedDW,Godefroid2008AutomatedWF} is a set of techniques that generate a variety of unexpected and different inputs for a program, executing and recording them if they cover parts of the program that have not yet been tested, result in crashes or hangs, or identify any unexpected behavior. 
The most promising feedback-based fuzzers are Grey-Box Mutational Fuzzers, such as \textbf{AFL} (``American Fuzzy Lop'') \cite{AFL}, which rely on light binary instrumentation and genetic algorithms to iteratively mutate inputs to increase code coverage. AFL has been broadly deployed, finding vulnerabilities in several popular software applications including Mozilla Firefox, Adobe Flash, and OpenSSL \cite{AFLBugs}. AFL selects a prior parent seed to sample, applies a set of mutation functions (number \textit{and} type) to it, and executes the program with the new child seed, adding the child to the queue if a new code path is discovered.

While there is a large body of work looking into improving grey-box fuzzing (and specifically AFL) with data-driven techniques, the work is concentrated on selecting the most promising parent inputs (or input subsequences) to mutate. For example, \citet{Lemieux2017FairFuzzTR} propose a new variant of AFL called FairFuzz, that focuses on mutating seeds that hit rare branches in the program control flow graph. Alternatively, \citet{Rajpal2017NotAB} use deep neural networks to learn which parts of an input correspond to magic bytes or boundaries that shouldn't be fuzzed (e.g. PDF headers), focusing fuzzing efforts on less constrained byte locations. One relatively less explored area is improvements from tuning the mutation process itself. For example, consider a simple program that performs carry addition of two command line inputs; in this case, any mutation operators that insert strings or non-numeric types will result in inputs that trigger top-level, error-handling code. Instead, we want to prioritize mutators that result in interesting numeric values, to create inputs that penetrate deep into the program.

In this work, we show how to learn a mutation operator distribution for generating new children that significantly improves upon AFL -- and this simple process gives larger gains than state-of-the-art parent selection techniques. We reformulate AFL's fuzzing loop as a multi-armed bandit problem, where each arm corresponds to a different mutation operator, and show that by using Thompson sampling \cite{pmlr-v23-agrawal12}, we can learn a better distribution over operators adaptively, \textit{during} the course of fuzzing an individual program, with drastically better results. Our results on the DARPA Cyber Grand Challenge Binaries \cite{DARPA}, a set of programs developed by humans and riddled with natural bugs show that our Thompson-sampling approach results in significantly more code coverage and more crashes found than other state-of-the-art fuzzers, including AFL version 2.52b, and FairFuzz \citep{Lemieux2017FairFuzzTR}. For completeness, we also evaluate on the LAVA-M synthetic bug dataset \cite{DolanGavitt2016LAVALA} where we show less conclusive results, possibly due to the small sample size (the set consists of only 4 programs), and the synthetic nature of the bugs. Taken together, our experiments on over 75 unique binaries show that tuning the mutator distribution adaptively yields significant benefits over existing fuzzers like AFL and FairFuzz.

\section{Fuzzing Overview}

\begin{table}[t]
\centering
\begin{tabular}{@{}lll@{}}
\toprule
Mutation Operation & Granularity        & Notes    \\ \midrule
Bitflips           & bit                & Flip single bit  \\
Interesting Values & byte, word, dword  & NULL, -1, 0, etc.\\
Addition           & byte, word, dword  & Add random value\\
Subtraction        & byte, word, dword  & Subtract random value\\
Random Value       & byte(s)            & Insert random value\\
Deletion           & byte(s)            & Delete from parent\\
Cloning            & byte (unbound)        & Clone/add from parent\\
Overwrite          & byte (unbound)          & Replace with random\\
Extra Overwrite    & byte (unbound)           & Extras: strings scraped\\
Extra Insertion    & byte (unbound)         & \qquad \quad from binary\\ \bottomrule
\end{tabular}

\vspace*{1mm}
\textbf{}\caption{List of mutation operators utilized by AFL.}
\label{tbl:afl-mutations}
\end{table}

Fuzzers are categorized by the amount of transparency into the program under test that they require. \textbf{Black-Box} fuzzers \cite{Woo2013SchedulingBM,Gascon2015PulsarSB,Radamsa,ZZUF} operate by generating a large number of random inputs very quickly without any information on the code being tested. The simplest type of black-box fuzzers are random fuzzers, such as Radamsa \cite{Radamsa}, and ZZUF \cite{ZZUF}. These work by sampling string lengths, then sampling characters or bits to generate uniformly at random. These types of fuzzers run extremely quickly, and are commonly used to fuzz tools for checking that user-supplied strings meet certain requirements (e.g. password checkers). While they can quickly find shallow bugs at the top levels of the programs under test, they take an extremely long time to find deeply nested bugs.

At the other extreme of program transparency, \textbf{White-Box} fuzzers \cite{Ganesh2009TaintbasedDW} and symbolic execution tools \cite{Li2013SteeringSE,Ma2011DirectedSE,Baldoni2018ASO,Pasareanu2011SymbolicEW,Wang2017AngrT,David2016BINSECSEAD} like KLEE \cite{Cadar2008KLEEUA} and SAGE \cite{Godefroid2012SAGEWF} symbolically manipulate the source code to deterministically discover input sequences that explore every code path. As they are able to walk the entire source code tree, they can discover bugs in nested code. These tools operate on intermediate representations of code like the LLVM IR, or assume high-level source code information. Such tools use constraint solvers to explicitly solve for inputs that reach certain branching points in code, and as a result degrade as programs grow large \cite{Krishnamoorthy2010TacklingTP,Baldoni2018ASO}. 

\subsection*{Grey-Box Fuzzing}
Between these two extremes are \textbf{Grey-Box} fuzzers \cite{Rawat2017VUzzerAE,Stephens2016DrillerAF,AFL,Lemieux2017FairFuzzTR,Bhme2016CoveragebasedGF,Li2017SteelixPB}. These fuzzers assume lightweight instrumentation can be injected into programs or binaries to determine coverage. This allows fuzzers to recognize when they have found inputs that exercise new code paths. While there are different types of grey-box fuzzers, by far the most effective are mutational fuzzers, which use genetic algorithms to ``mutate'' a parent input into a new child input, allowing one to reason about how similar inputs affect program execution.

A general sketch for mutational grey-box fuzzing can be found in Algorithm~\ref{alg:greybox}. While the algorithm depicted in the figure is specific to AFL, the preeminent mutational grey-box fuzzer, the general algorithm takes the same form across all mutational fuzzers, with different heuristics driving the key functions ({\tt pick_input}, {\tt sample_mutation}, etc.). As a brief summary, these fuzzers work by maintaining a Queue of inputs that exercise different code paths. The fuzzing process is then reduced to the two loops on lines 6 and 8 of Algorithm \ref{alg:greybox}, with the outer loop (line 6) responsible for picking an input and a corresponding energy value from this Queue ({\tt pick_input}). One can think of the energy as the amount of time to spend fuzzing a given parent input (the number of iterations to run the inner loop from Algorithm \ref{alg:greybox}). The inner loop (line 8) is then responsible for creating new child inputs from the parent, by applying a random number ({\tt sample_num_mutations}) of specific mutation functions ({\tt sample_mutation}). These mutation functions include random bit-flips, random byte-flips, adding ``magic'' bytes, arithmetic operations at the bit level, etc., and are applied to the chosen input at a particular byte offset ({\tt sample_mutation_site}) to create a new child candidate. A comprehensive table of the mutation operators used by AFL can be found in Table \ref{tbl:afl-mutations}. This child candidate is then executed, and if it exercises an code path that is different from those exercised by the inputs already logged, it is added to the queue. 

\subsection{AFL}

One of the most popular and widely-adopted grey-box fuzzers is ``American Fuzzy Lop'' \cite{AFL} or AFL. Since we build our approach on top of AFL, it is worthwhile to understand its strengths as well as its limitations. The core of AFL's algorithm is detailed in Algorithm \ref{alg:greybox}; however, while this highlights the main loop behind AFL, it fails to highlight some key details. 

The first is AFL's feedback-driven coverage mechanism. AFL is efficient and intelligent; during fuzzing, it can execute hundreds of inputs per second, traversing a large amount of the program input space all while optimizing for code coverage. It does this by injecting lightweight instrumentation into the program under test, to track the path an input takes through the program. It works as follows: AFL first assigns each basic block in the program a random ID. These block IDs then compute edge IDs (identifying transitions from one basic block to another, or edges in the program control flow graph), allowing AFL to track paths. Specifically, AFL stores a bitmap that maps transitions to rough ``hit-counts'' (number of times a given transition has been observed for a single input). Any input that observes a transition that has not yet been seen is deemed ``interesting,'' and is added to the Queue (line 14 of Algorithm \ref{alg:greybox}). This tracking instrumentation can either be injected at compile time (by compiling programs with the afl-gcc or afl-clang compilers that ship with AFL), or at runtime via the QEMU emulator, allowing AFL to fuzz closed-source binaries. In our experiments, we work exclusively on binaries, utilizing this QEMU mode.

The second detail pertains to AFL's implementation of \texttt{pick_input}, or how to choose which queue entry to fuzz next. Note that on line 5 of Algorithm \ref{alg:greybox}, the \texttt{pick_input} function consumes the queue as an argument, and returns two outputs, the ``parent'', or the actual queue entry to mutate into new children, as well as an ``energy'' value, corresponding to how long to spend fuzzing (iterations of the inner loop starting on line 6) the given parent. AFL implements this function based solely on heuristics: namely, AFL defines a \texttt{calculate_score} function that assigns energy for each queue entry, based on the following factors: coverage (prioritize inputs that cover more of the program), execution time (prioritize inputs that execute faster), and discovery time (prioritize inputs discovered later). 

The final detail pertains to AFL's modes of operation. The fuzzer is currently shipped with two distinct modes, \textbf{AFL} and \textbf{FidgetyAFL}, as they are referred to by AFL's developer, as well as in prior work \cite{FidgetyAFL,Lemieux2017FairFuzzTR}. While the core of both versions of AFL is the loop depicted in Algorithm \ref{alg:greybox}, the key difference is that standard AFL runs a series of controlled, \textit{deterministic} mutations on each parent input, prior to starting the main fuzzing loop. These deterministic mutations operate at every bit/byte of each input in the queue sequentially, and include the following operations \cite{AFLTechnical}:\begin{itemize}
	\item Bit flips, with varying lengths and offsets.
    \item Arithmetic operations (add/subtract small integers).
    \item Insertion of ``interesting'' integers (from a finite set including 0, 1, MAX/MIN_INT, etc).
\end{itemize}
The goal of these deterministic mutations is to create test cases that capture fixed boundaries in input space. However, prior work \cite{Bhme2017DirectedGF,Bhme2016CoveragebasedGF}, as well as the lead AFL developer \cite{FidgetyAFL} note that in many programs, one can obtain more coverage and discover more crashes by omitting this deterministic stage. Indeed, AFL without this deterministic stage is exactly FidgetyAFL, or the loop depicted in Algorithm \ref{alg:greybox} (AFL run with the ``-d'' parameter). In our experiments, we utilize both AFL and FidgetyAFL as baselines (FidgetyAFL is almost always the stronger fuzzer of the two).

Both versions of AFL ultimately enter the loop in Algorithm \ref{alg:greybox}. 
AFL chooses a number of mutations to apply (\texttt{sample_mutations()} on line 8 of Algorithm \ref{alg:greybox}) uniformly at random from the set of powers of 2 between $2^1$ to $2^7$. The mutation operators (\texttt{sample_mutation()} on line 9 of Algorithm \ref{alg:greybox}) are also picked uniformly at random from the set of 16 operators in Table \ref{tbl:afl-mutations}. The site the given mutation is applied (\texttt{sample_mutation_site()} on line 10 of Algorithm \ref{alg:greybox}) is also chosen uniformly at random over the bytes in the parent.


\section{Related Work}

There is a large body of work focused on learning better heuristics for picking inputs to fuzz (\texttt{pick_input} on line 5 of Algorithm \ref{alg:greybox}). Both \citet{Bhme2017DirectedGF} and \citet{Rawat2017VUzzerAE} learn to pick inputs in the queue, in addition to energy values, to maximize the probability of hitting a desired part of the program, or exercise a desired code path (if one wants to fuzz a certain function in the program, they should prioritize inputs in the queue that exercise code paths that utilize the given function). Both these approaches use lightweight learning (in the case of \citet{Bhme2017DirectedGF}, a learned power schedule) to build on top of AFL, and show strong results. Furthermore, \citet{Bhme2016CoveragebasedGF} introduce a set of algorithms for optimizing fuzzing, based on a probabilistic model of coverage. They introduce AFLFast, a AFL-variant that introduces new search strategies and power schedules to maximize code coverage. Specifically, the authors are able to show that their approach beats AFL and FidgetyAFL \cite{FidgetyAFL} on the GNU binutils, a set of standard Linux programs.

More recently, \citet{Lemieux2017FairFuzzTR} introduce FairFuzz, another technique to optimize \texttt{pick_input}. Specifically,the authors work under the assumption that the best way to find crashes in programs and improve code coverage is to prioritize inputs that exercise rare branches. By focusing on fuzzing these inputs, FairFuzz generates many new inputs that hit the extremes of a program, finding a multitude of bugs. FairFuzz also optimizes the selection of mutation sites ($\texttt{pick_mutation_site}$) by tracking which bytes in the input lead to changes in control flow. They use this to prevent fuzzers from touching volatile regions of the input buffer. The results in \cite{Lemieux2017FairFuzzTR} show that FairFuzz is significantly more effective than current versions of AFL, FidgetyAFL, and AFLFast -- making it the most effective grey-box fuzzer released at the time of this work. Related, \citet{Rajpal2017NotAB} use deep neural networks to select mutation sites (\texttt{pick_mutation_site}), in order to prevent fuzzers from creating trivial inputs that fail error handling code or syntax checks. Specifically, the authors show that they can effectively fuzz PDF parsers, and generate non-trivial PDFs. 

All the above fuzzers leave the distribution over mutation operators untouched, defaulting to the uniform distribution that comes with AFL. One of the few works that examines the mutation distribution is that by \citet{Bttinger2018DeepRF} that frames grey-box fuzzing as a reinforcement learning problem, and use Q-Learning \cite{Watkins1992Qlearning} to learn a policy for choosing mutation operators. 
While this approach is quite interesting, they only evaluate on a single program (a PDF parser), with a single queue entry that remains fixed as opposed adjusting the queue during fuzzing. Moreover, their baseline is a random fuzzer not AFL. Here, we compare against state of the art fuzzers on a standard set of programs.

\section{Approach}

In this section, we present the details of our approach for estimating distributions over grey-box fuzzing mutators. Section \ref{sec:motivation} presents a couple of motivating examples, Section \ref{sec:empirical} specifies a method for learning a better stationary distribution over mutators (to verify that there do exist better mutator distributions than the uniform distribution used with AFL), and Section \ref{sec:thompson} presents our adaptive, Thompson Sampling method. 

\subsection{Motivating Examples}
\label{sec:motivation}

To understand the effect the mutator distribution has on fuzzing efficiency, we present two motivating examples. Both the examples we walk through are part of the DARPA Cyber Grand Challenge (CGC) dataset \cite{DARPA}. They have different functionality and varying levels of complexity, as one would expect of real-world, human-written programs.

\subsubsection{ASCII Content Server}

The first example can be found in Listing \ref{lst:ascii}, a simplified excerpt of the C program \texttt{ASCII_Content_Server}, which demonstrates top-level functionality for a simple text-based HTTP web server. The program supports functionality for starting up a web session, in which a user (over STDIN) can issue a series of commands for creating web pages (``SEND''), looking up, interacting, and visualizing pages (``QUERY'', ``INTERACT'', ``VISUALIZE''), and issuing API calls (``REQUEST''). Sessions can consist of arbitrarily many of these commands, as they are all specified over STDIN. However, note that each command begins with the keyword (as a string) of the given type, followed by the data for the given command (e.g. ``REQUEST <DATA>''). 

Consider the mutation operators listed in Table \ref{tbl:afl-mutations} in the context of this program. Many of these mutators operate randomly, at the bit level (e.g. Bitflips, Addition/Subtraction). Especially because the site of these mutation operators is also uniformly distributed, these bit-level operators prove to be less effective at producing good inputs for the program, as they have a good chance of colliding with the keyword strings (e.g. ``REQUEST'') that dictate the command. As such, the corresponding child inputs will, with high likelihood, end up malformed, triggering the default error handling code on line 26. This prevents the fuzzer from exploring deeply into the program, as with a uniform distribution over mutators, interesting inputs will only be produced with low likelihood.

Instead, to increase the likelihood of generating interesting inputs, reconsider the mutators in Table \ref{tbl:afl-mutations}. Namely, consider the mutators for Inserting/Overwriting ``Extras'' (in this case, the keyword strings ``REQUEST,'' ``QUERY,'' etc.), and for Cloning bytes from the parent seed. Both these mutators are significantly more likely to produce interesting inputs, as they are more likely to result in well-formed inputs to pass the syntactic check in the \texttt{ReceiveCommand} function. Indeed, especially later in the fuzzing process, one might expect the Clone mutator to be even more effective, as it copies longer and longer sessions of commands into child inputs, allowing for more complex sessions that could trigger non-trivial bugs. 

\begin{listing}
\begin{minted}{c}
// Runs the ASCII Content Server
int main(void) {
  // Initialize server
  InitializeTree();

  // Respond to commands
  Command command, int more = {}, 1;
  while (more) {
    ReceiveCommand(&command, &more);
    HandleCommand(&command);
  }
  return 0;
}

// Receives and parses an incoming command
int ReceiveCommand(Command *command, int *more) {
  char buffer[64], size_t bytes_received;
  read_until(buffer, ':', sizeof(buffer));

  switch (buffer) {
    case "REQUEST": command->c = REQUEST; break;
    case "QUERY": command->c = QUERY; break;
    case "SEND": command->c = SEND; break;
    case "VISUALIZE": command->c = VISUALIZE; break;
    case "INTERACT": command->c = INTERACT; break;
    default: more = 0; return -1;
  }
  parse_data(command, read_rest(), more);
  return 0;
}

// Handle a command, and update/read from the DOM
void HandleCommand(Command *command) {
  switch (command->c) {
    case REQUEST: ServePage(Lookup(command->n));
    ...
    
\end{minted}
\caption{ASCII_Content_Server.c \medskip \\  Simplified extract capturing Program ASCII_Content_Server from the DARPA CGC binaries. Provides functionality for a simple, text-based HTTP-like web server.}
\label{lst:ascii}
\end{listing}

\subsubsection{ASL6 Parser}
\begin{listing}
\begin{minted}{c}
#define PRIM_MASK 0x20
#define CLASS_MASK 0xC0
#define TAG_MASK 0x1F

// Runs the Parser
int main(void) {
  unsigned sz, uint8_t b[MAX_SIZE];
  receive(STDIN, &sz, sizeof(sz));
  read_exactly(STDIN, b, sz);
  pretty_print(decode(b, 0, (unsigned) b, 
                     (unsigned) b + sz));
}

// Decode the T-L-V format str into an element
element *decode(uint8_t *b, unsigned depth, 
                unsigned st, unsigned sp) {
  element *e = malloc(sizeof(element));

  // Parse out class and tag (message)
  e->primitive = (b[0] & PRIM_MASK) == 0;
  e->cls = (b[0] & CLASS_MASK) >> 6;
  e->tag = (*b & TAG_MASK);

  // Parse out additional data
  int bread = parse_length(b + 1, e);
  e->data = b + bread;

  // Recursively Parse
  if (!e->primitive) {
    uint8_t cur = e->data;
    while (1) {
      sub = _decode(cur, depth + 1, st, sp);
      append_sub(e, sub) < 0);
      cur = sub->data + sub->length;
      if (cur >= e->data + e->length) break;
    }
  }
  return e;
}
\end{minted}
\caption{ASL6parse.c \medskip \\ Simplified extract capturing Program ASL6parse from the DARPA CGC binaries. Provides functionality for parsing and pretty-print a type-length-value formatted string (similar to ASN.1 encoding).}
\label{lst:asl}
\end{listing}

The second example can be found in Listing \ref{lst:asl}, a simplified excerpt of the C program \texttt{ASL6parse}. \texttt{ASL6parse} provides functionality for a recursive parse for the ASL6 formatting standard, a type-length-value format akin to ASN.1 encoding. The core of the functionality provided by this program involves reading arbitrarily long ASL6-encoded strings over STDIN, decoding them into a more easily human-readable format, and then pretty-printing the results to screen. Each ASL6 string takes the format of a series of messages, where each message consists of three values: 1) a type or class, 2) a length value, specifying how many characters the given value has (to facilitate reading), and 3) the string value, which consists of either a primitive, or one or more nested messages. 

Unlike the prior program, there is no reliance in \texttt{ASL6parse} on any keywords, or specific string values; indeed, each tag/class is found by taking the bitwise-and of the initial bits, with a constant ``TAG_MASK.'' As such, all that is necessary to explore deeper into the program is to generate more complex strings that loosely follow the rather relaxes ASL6 format. If we consider the mutators from Table \ref{tbl:afl-mutations}, we see that those that prioritize random insertion, coupled with those arithmetic operators (for bumping the length values) are more likely to produce interesting inputs for the program than those that overwrite/insert string values -- an effective distribution would prioritize the former operators, and de-prioritize the latter.

It is interesting to note that across two different programs, there is a great need for drastically different distributions over mutators (note how the most effective mutators for \texttt{ASCII_Content_Server} are the \textit{least} effective for \texttt{ASL6parse}). It is clear that a solution that learns a non-uniform distribution over mutators would improve the likelihood of generating inputs that penetrate deeper into programs. Furthermore, this solution needs to be adaptive, able to learn which operators are more effective than others on a per-program basis.

\subsection{Estimating Distributions Empirically}
\label{sec:empirical}

Our hypothesis is that best way to optimize fuzzing is to learn mutator distributions adaptively. However, it is necessary to assess whether there exist better fixed mutator distributions than the uniform distribution utilized by AFL, to get a sense of exactly how inefficient AFL's current heuristics are. To do this, we outline an approach that finds a fixed distribution by analyzing and extrapolating from a set of full fuzzing runs on a set of programs that are unrelated to those we test on. Intuitively, the goal here is to use data collected on unrelated programs to learn a new mutator distribution that performs better than AFL out-of-the-box. While this empirically estimated stationary distribution is not optimal for an individual program (as it is not tuned adaptively), our approach for learning it has two key benefits: 1) it lets us get a sense of AFL's current inefficiency while providing a strong point of comparison for our adaptive Thompson Sampling strategy, and 2) it allows us to provide intuition as to how to use data to inform the choice of mutator distributions, before delving into the adaptive, Thompson sampling approach detailed in the next section (Section \ref{sec:thompson}).


Our approach for estimating a better stationary distribution over mutators is straightforward. Recall that, as with most fuzzers, the metric we are optimizing for is code coverage -- we wish to maximize the number of code paths we discover (again, other metrics like crashes or hangs are too sparse and random to serve as a meaningful success signal). Our approach is simple: the best indication of mutation operators that will be successful in the future are mutation operators that have been successful in the past. 

Consider the following: Let $c_k$ correspond to the number of times mutation operator $k$ was utilized in creating a successful input. A successful input is a child mutant that explores a new code path. We can track this quantity by adding some lightweight code to AFL to log mutation counts for every mutant. With these aggregate counts, we can estimate a new distribution as follows:
\begin{align}
	p_k = \dfrac{c_k}{\sum_{k'=1}^K c_{k'}}
\end{align}
Where $p_k$ is the probability of choosing mutation operator $k$, from the $K$ possible mutations. Note that the role of the denominator here is solely to make sure the probabilities $p_k$ sum to 1. This estimation of the probability parameters based on success counts is exactly Maximum Likelihood estimation. Results capturing the efficacy of this approach can be found in Section \ref{sec:empres}.

\subsection{Thompson Sampling}
\label{sec:thompson}

In this section, we derive our approach for learning the distribution over mutation operators adaptively. The intuition of this approach is similar to that of the prior section; we should prioritize mutation operators that have been successful in the past. However, unlike before, where we estimated our distribution at the end of many complete fuzzing runs on a large suite of programs, here we don't have the same amount of data, nor the same measure of confidence because we are working on an individual program, where we start estimating a new distribution after just a short period of fuzzing. Instead, we need an approach that allows us to factor in our confidence about our estimates, starting off cautious, then becoming more sure as we gather more data. One approach that gives us this behavior is Thompson Sampling \cite{pmlr-v23-agrawal12}.

Thompson Sampling is an approach for the classic Multi-Armed Bandit problem. In a Multi-Armed Bandit, the setup is as follows: there are $K$ slot machines (``bandits''), each with an unknown probability of paying off. The problem proceeds in rounds, and in each round, we can select one of the $K$ bandits to play. Depending on the family of distributions utilized for each bandit, we observe some reward. For example, in the Bernoulli bandit setting, the payoff for each bandit is drawn from a Bernoulli distribution, parameterized by $\theta_k$ for a given bandit $k$ -- bandit $k$ will payout 1 with probability $\theta_k$, and 0 with probability $1 - \theta_k$. 

Intuitively, each $\theta_k$ corresponds to a probability of success of playing a given bandit. To maximize reward, then, it is best to play the arm with the highest success probability. In other words, if we had full information, and knew the parameters $\theta_1, \theta_2, \ldots \theta_K$ for each bandit arm, the optimal strategy would be to always play $k_{\text{best}} = \argmax_k \theta_k$. However, we don't have this information. As a result, we need to trade off between trying  various bandits to estimate their probabilities of payoff, and committing to the bandit that seems the best so far. 

We set up our problem of adaptively estimating a distribution over mutators as a Multi-Armed Bandit problem as follows: Let each mutation operator $1, 2 \ldots K$ (the 16 operators detailed in Table \ref{tbl:afl-mutations}) be a different bandit. For each mutation operator, let $\theta_k$ be the likelihood that the corresponding operator is used to generate a successful input (an input that increases code coverage). Finally, instead of picking the ``best'' arm $k$, like in the standard Multi-Armed Bandit section, set the distribution over mutation operators as:
\begin{align}
	p_k = \dfrac{\theta_k}{\sum_{k'=1}^K \theta_{k'}}
\end{align}
where $p_k$ is the probability of choosing operator $k$ during fuzzing. 

With this, all that's left is to work out the details for handling the exploration-exploitation tradeoff, and best learn each $\theta_k$.

\subsubsection{Exploration-Exploitation}

Thompson Sampling frames the exploration-exploitation trade-off as a Bayesian posterior estimation. Returning to the traditional Multi-Armed Bandit setting, we start with a prior distribution on each of the $K$ bandits $\pi(\theta_1), \pi(\theta_2), \ldots \pi(\theta_K)$. This prior is initially set to reflect the starting belief of how the parameters of each bandit are set (in our case, our prior is the AFL distribution over operators -- namely, a uniform distribution). Then, let $D_t$ be the data we have collected in the first $t$ rounds of play, which we can represent by the pairs $(n_{10}, n_{11}), (n_{20}, n_{21}), \ldots (n_{K0}, n_{K1})$, where $n_{k1}$ is the number of times the given arm resulted in a payoff of 1, and where $n_{k0}$ is the number of times the given arm resulted in a payoff of 0. So if we look at all the data we collected by round $t$, across all arms, we see that $\sum_{k=1}^K (n_{k0} + n_{k1}) = t$. 

Suppose we've just completed round $t$, observing data $D_t$. We want to combine our data $D_t$ with our prior knowledge, and collect the posterior distribution for each arm. These posterior distributions $\pi(\theta_1 \mid D_t), \pi(\theta_2 \mid D_t), \ldots \pi(\theta_K \mid D_t)$ represent our rationally updated beliefs about the values of $\theta_1, \theta_2, \ldots \theta_K$. Once we have these posterior distributions, the question remains: which bandit should we play at the $t + 1$'st round? A greedy approach would be to play the bandit that has the highest probability of being the best (the bandit $k$ with the highest $\theta_k$) -- yet this is all exploitation, and no exploration. The Thompson Sampling approach instead compromises in an interesting way: For each bandit $k$, sample $\hat{\theta_k} \sim \pi(\theta_k \mid D_t)$. Then, play bandit $\tilde{k} = \argmax_{k} \hat{\theta_k}$. Note that this $\tilde{k}$ is a sample from the posterior distribution over which bandit has the highest probability of paying off. As we become more confident about which is the best bandit (as we collect more data!), we'll choose this bandit more frequently -- which is exactly what we wanted. Next, we work out the details for computing the posterior distributions.

\subsubsection{The Beta/Bernoulli Model}

In our setup, we utilize Bernoulli distributions as the payoff distributions (we want 0/1 reward). We choose our prior distributions from the Beta family, for it's conjugacy to the Bernoulli distribution:
\begin{align*}
\theta &\sim \text{Beta}(\alpha, \beta) \\
\pi(\theta) &\propto \theta^{\alpha - 1} (1 - \theta)^{\beta - 1}
\end{align*}
where $\pi(\theta)$ is supported on the interval $(0, 1)$, and $\alpha, \beta$ are positive real valued parameters. 

Since the $K$ bandits are independent (by assumption), we can update their posterior distributions independently. For each $\theta_k$, the prior is taken to be $\theta_k \sim \text{Beta}(\alpha_{k}, \beta_{k})$. The likelihood function for $\theta_k$ is $\Pr(n_{k0}, n_{k1} \mid \theta_k) = \theta_k^{n_{k1}} (1 - \theta_k)^{n_{k0}}$. We calculate the posterior density to be:
\begin{align*}
	\Pr(\theta_k \mid D_t) &\propto \Pr(\theta_k) \Pr(D_t \mid \theta_k) \\
    &\propto \theta_k^{\alpha_k - 1} (1 - \theta_k)^{\beta_k - 1} \times \theta_k^{n_{k1}} (1 - \theta_k)^{n_{k0}} \\
    &= \theta_k^{\alpha_k - 1 + n_{k1}} (1 - \theta_k)^{\beta_k - 1 + n_{k0}}
\end{align*}
Thus $\theta_k \mid D_t \sim \text{Beta}(\alpha_k + n_{k1}, \beta_k + n_{k0})$. Using this expression, we can update our posterior distributions with the data observed after each round, and choose the next bandit (next mutator) using the sampling approach described above.

The above process is straightforward to implement in our fuzzing setup. To update our posteriors, we need only track how many times each operator $k$ was involved in generating a successful ($n_{k1}$) or unsuccessful ($n_{k0}$) input. 

\begin{table*}[t]
\resizebox{.7\linewidth}{!}{
\begin{tabular}{@{}lccccc@{}}
\toprule
                  & 30 Min        & 60 Min        & 90 Min        & 120 Min       & Wins / FidgetyAFL \\ \midrule
AFL               & 0.58 $\pm$ .03  & 0.60 $\pm$ 0.03 & 0.60 $\pm$ 0.03 & 0.60 $\pm$ 0.03 & 6                 \\
FidgetyAFL        & 0.81 $\pm$ .02  & 0.82 $\pm$ 0.02 & 0.82 $\pm$ 0.02 & 0.82 $\pm$ 0.02 & ---               \\
FidgetyAFL, n=1   & 0.78 $\pm$ 0.03 & 0.79 $\pm$ 0.03 & 0.79 $\pm$ 0.03 & 0.80 $\pm$ 0.03 & 34                \\
FidgetyAFL, n=2   & 0.83 $\pm$ 0.03 & 0.84 $\pm$ 0.02 & 0.84 $\pm$ 0.02 & 0.84 $\pm$ 0.03 & 42                \\
\textbf{FidgetyAFL, n=4}   & \textbf{0.84 $\pm$ 0.02} &\textbf{ 0.85 $\pm$ 0.02} & \textbf{0.84 $\pm$ 0.02} & 0\textbf{.85 $\pm$ 0.02} & \textbf{44}                \\
FidgetyAFL, n=8   & 0.83 $\pm$ 0.02 & 0.82 $\pm$ 0.02 & 0.83 $\pm$ 0.02 & 0.84 $\pm$ 0.02 & 41                \\
FidgetyAFL, n=16  & 0.78 $\pm$ 0.03 & 0.80 $\pm$ 0.02 & 0.79 $\pm$ 0.02 & 0.79 $\pm$ 0.02 & 31                \\
FidgetyAFL, n=32  & 0.75 $\pm$ 0.02 & 0.75 $\pm$ 0.02 & 0.74 $\pm$ 0.02 & 0.75 $\pm$ 0.02 & 24                \\
FidgetyAFL, n=64  & 0.71 $\pm$ 0.03 & 0.71 $\pm$ 0.03 & 0.70 $\pm$ 0.03 & 0.71 $\pm$ 0.03 & 20                \\
FidgetyAFL, n=128 & 0.67 $\pm$ 0.03 & 0.67 $\pm$ 0.03 & 0.67 $\pm$ 0.03 & 0.67 $\pm$ 0.03 & 14                \\ \bottomrule
\end{tabular}}
\caption{Results for \texttt{sample_num_mutations} experiments. Numbers in the first four columns correspond to mean and standard error of relative coverage statistics, across the 75 training programs, while the ``Wins / FidgetyAFL'' column corresponds to how many times the given strategy discovered more code paths than FidgetyAFL, the best baseline.}
\label{tab:nstack}
\end{table*}

\subsubsection{Credit Assignment}
\label{sec:credit}

There is one slight complication; in a standard Multi-Armed Bandit setting, each pull of a bandit arm immediately translates to the outcome -- either 0 or 1 reward. In this way, we can exactly assign the score to the corresponding bandit. However, in our scenario, we sample an arbitrary number of mutations to apply before seeing whether or not an input results in additional code coverage (see line 8 of Algorithm \ref{alg:greybox}). This means we can possibly see many different mutation operators (up to 128, given AFL's default parameters) involved in the creation of a new input. With this, it is hard to assign the proper reward to each mutation involved, as it's unclear exactly which mutation operator led to the success (or failure). Do we assign each mutation operator involved a score of 1 if the final outcome is successful? What happens when the same mutation operation shows up multiple times?

This is known as the credit assignment problem, and it slightly muddles the approach outlined above. However, we find that by making one slightly minor change, we can get around this problem. We hypothesize that the high variance in the number of mutations is the problematic piece; if instead, AFL chose a smaller, constant number of mutations (i.e. replace the loop iteration count on line 8 of Algorithm \ref{alg:greybox} with a small constant), the effect of credit assignment would be diminished. To do this, we return to the training program set utilized in Section \ref{sec:empirical}. Much like we learned a good stationary mutator distribution from data, we can similarly learn a good small constant, for the number of mutation operators to apply. If the corresponding AFL with a fixed number of mutations performs as well as or better than AFL (or FidgetyAFL), then it might serve as a good base for our Thompson Sampling implementation. The experimental setup and results for estimating a fixed number of mutations to apply can be found in Section \ref{sec:num-stacks}.

\section{Datasets}

We utilize two program datasets to highlight the efficacy of our approach: the DARPA Cyber Grand Challenge binaries \cite{DARPA} and the LAVA-M binaries \cite{DolanGavitt2016LAVALA}.

\subsection{DARPA Cyber Grand Challenge}

\begin{table*}[t]
\begin{tabular}{@{}lccccccc@{}}
\toprule
                  & 6 hr          & 12 hr         & 18 hr         & 24 hr         & Crashes & Wins / FidgetyAFL & Wins / All \\ \midrule
AFL               & 0.64 $\pm$ 0.03 & 0.63 $\pm$ 0.03 & 0.63 $\pm$ 0.03 & 0.63 $\pm$ 0.03 & 554     & 18                & 4          \\
FidgetyAFL        & 0.84 $\pm$ 0.02 & 0.84 $\pm$ 0.02 & 0.85 $\pm$ 0.02 & 0.84 $\pm$ 0.02 & 780     & ---               & 14         \\
Empirical    & 0.85 $\pm$ 0.02 & 0.86 $\pm$ 0.02 & 0.86 $\pm$ 0.02 & 0.87 $\pm$ 0.02 & 766     & 41                & 5          \\
\textbf{Thompson} & \textbf{0.91 $\pm$ 0.02} & \textbf{0.92 $\pm$ 0.02} & \textbf{0.92 $\pm$ 0.02} & \textbf{0.93 $\pm$ 0.02} & \textbf{1336}    & \textbf{52}                & \textbf{47}         \\ \bottomrule
\end{tabular}
\caption{Results on the 75 DARPA CGC Test binaries. Wins are reported excluding ties. Thompson sampling has the highest Relative Coverage, and more total unique crashes (unique code paths leading to crash) than any other strategy.}
\label{tbl:darpa-res}
\end{table*}

The DARPA Cyber Grand Challenge binaries are a set of 200 programs released by DARPA as part of an open challenge to create tools for automatically finding, verifying, and patching bugs autonomously. The challenge took place in 2016, and many of the competing teams utilized grey-box fuzzers like AFL, in addition to other tools for symbolic execution. 

Each of the 200 binaries released as part of the challenge were written by humans, and each of the programs have widely different functionalities. Some binaries are simple, variants of simple programs for sorting or performing calculator operations, but many of the programs are arbitrarily complex (like the text-only HTTP server from Section \ref{sec:motivation}), spanning hundreds of lines. However, what all these binaries have in common is that each contains one or more bugs, documented by the program writers. These bugs are natural, in that they are programming errors that a human would actually make, and as such make this dataset the ideal test suite for benchmarking fuzzing applications. These bugs include simple errors like off-by-one errors, or use-after-frees, but many of the bugs are complex and hidden deep into programs, and can only be found through full exploration; as an example, consider the program \texttt{ASCII_Content_Server.c} from Section \ref{sec:motivation}. One of the bugs in this program (there are 3) is a null pointer dereference that only occurs after performing a ``VISUALIZE'' command after putting several pages of more than a certain size -- vulnerabilities like this that depend on several linked interactions are the common mode of the bugs in the CGC dataset, again making it an excellent test suite for benchmarking fuzzers.

While there are 200 binaries released as part of the DARPA CGC dataset, we only utilize the 150 (split into 75/75 train and test programs) that accept input over STDIN, as opposed to as command line arguments, or through files. This is for simplicity - to allow our version of AFL to interface with the remaining binaries, we would have needed to rewrite each program, or build a program-specific pre-processor/loader. However, we found the 150 programs to be sufficient to draw conclusions from our fuzzing results. Furthermore, while the original CGC binaries were released for a DARPA-specific OS called DECREE, we utilize ported versions of the binaries, compiled for x86 Linux OS. The full C code for each binary, and cross-platform compilation instructions can be found here: \url{https://github.com/trailofbits/cb-multios}.

\subsection{LAVA-M}
\label{sec:lava}

We also utilize the LAVA-M program dataset \cite{DolanGavitt2016LAVALA}, 4 programs taken from the GNU Coreutils suite, injected with thousands of synthetic bugs. This dataset has become popular in recent years, for benchmarking complex white-box fuzzers and symbolic execution tools, in addition to certain grey-box fuzzers (and some variants of AFL). The LAVA-M binaries also come with tools for the easy de-duplication of crashes, allowing us to report actual bugs discovered by our fuzzers. The synthetic bugs in the LAVA-M binaries are highly complex, placed at various deep points in the program (meaning that random inputs are highly unlikely to find them). 

Specifically, the LAVA-M dataset consists of the following four programs for GNU Coreutils version \texttt{8.24}: 1) \texttt{base64}, run with the \texttt{-d} (decode) option, 2) \texttt{md5sum}, run with the \texttt{-c} (check digests) option, 3) \texttt{uniq}, and 4) \texttt{who}. These programs were injected with 28, 44, 57, and 2265 bugs respectively. All bugs injected into the LAVA-M programs are of the following type: bugs that only trigger when a certain offset of the program input buffer matches a 4-byte ``magic'' (random) value (for more info see page 10 of \cite{DolanGavitt2016LAVALA}).

\section{Experimental Results}

In this section we provide details for each batch of experiments we run. Many of the following experiments consist of comparing many different strategies (i.e. Thompson Sampling vs. FidgetyAFL, vs. AFL). For many of our experiments, we evaluate against both AFL and FidgetyAFL as baselines -- in these cases, we treat FidgetyAFL as our best baseline, due to its significantly better performance. In addition to reporting aggregate counts on which strategy performs best on a per-program basis, we also report Relative Coverage, a metric which loosely provides a measure of how much better one strategy is relative to the others (as AFL doesn't instrument the program to give us proper basic block coverage statistics): Let $s$ correspond to a given strategy, $t$ the given time interval, and \textit{$\text{code-paths}_t(s)$} be the number of unique code-paths that strategy $s$ has discovered by time $t$. Then Relative Coverage \textit{$\text{rel-cov}$} for a single program can be computed as:
\begin{align*}
	\text{rel-cov}_t(s) &= \dfrac{\text{code-paths}_t(s)}{\max_{s'} [\text{code-paths}_t(s')]}
\end{align*}
or the ratio between the number of code paths strategy $s$ has found, and the maximum number of code paths across all strategies. We report the mean/standard error of this number across all programs. Furthermore, to get a better sense of the general story of each fuzzing run, we report statistics at each quartile of the total time (e.g. at 6, 12, 18, and 24 hours of a 24 hour run). Note that this is a \textit{relative} statistic, estimated at a single time interval; this means that this number may fluctuate for a given strategy across different time intervals, because the gap between two strategies may grow or diminish (this is the case in Table \ref{tbl:darpa-res}).  

All of the following experiments are conducted on machines running Ubuntu 14.04, with 12 GB of RAM and 2 cores. Prior to starting fuzzing, we scrape all string constants from the binary we are testing, to serve as dictionary entries during fuzzing.
\subsection{Estimating $\texttt{sample\_num\_mutations}$}
\label{sec:num-stacks}

\begin{figure*}
\includegraphics[width=.265\linewidth]{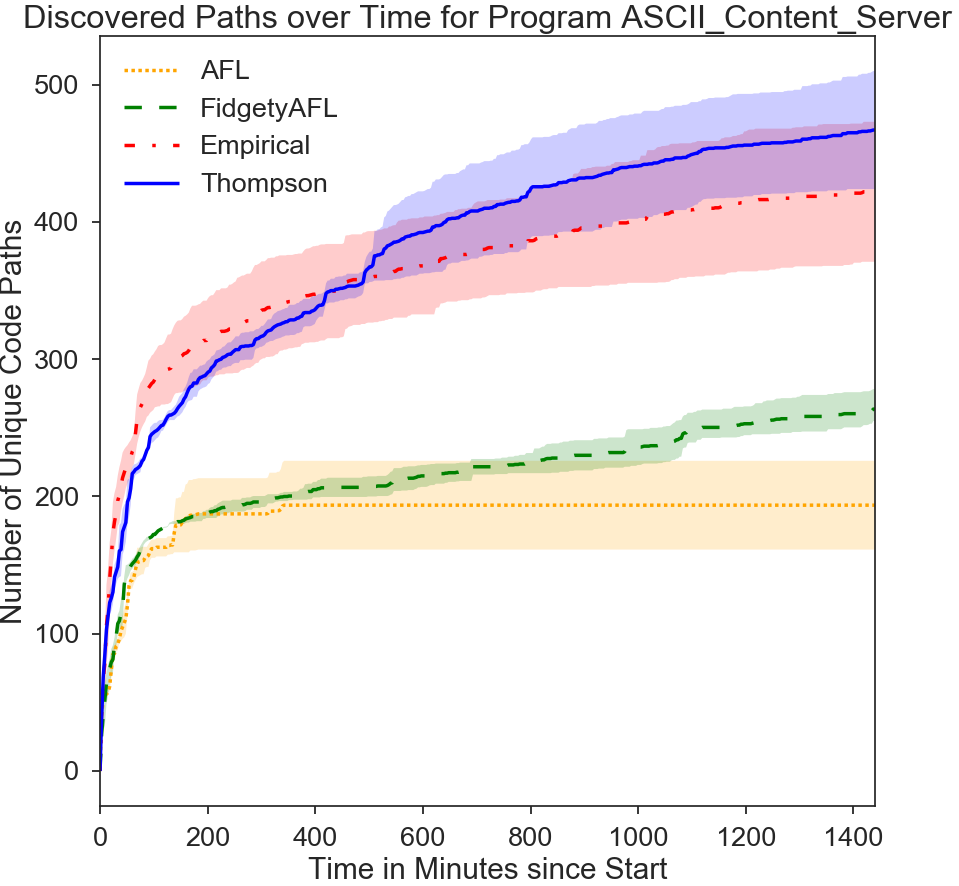}
\hfill
\includegraphics[width=.27\linewidth]{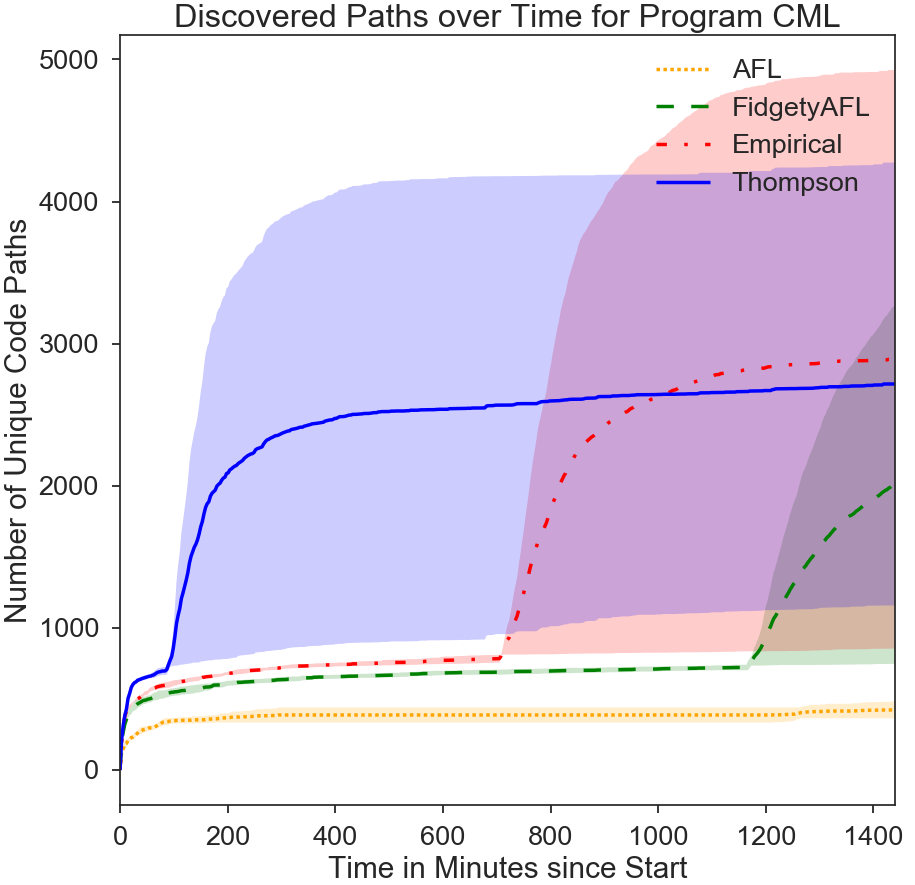}
\hfill
\includegraphics[width=.27\linewidth]{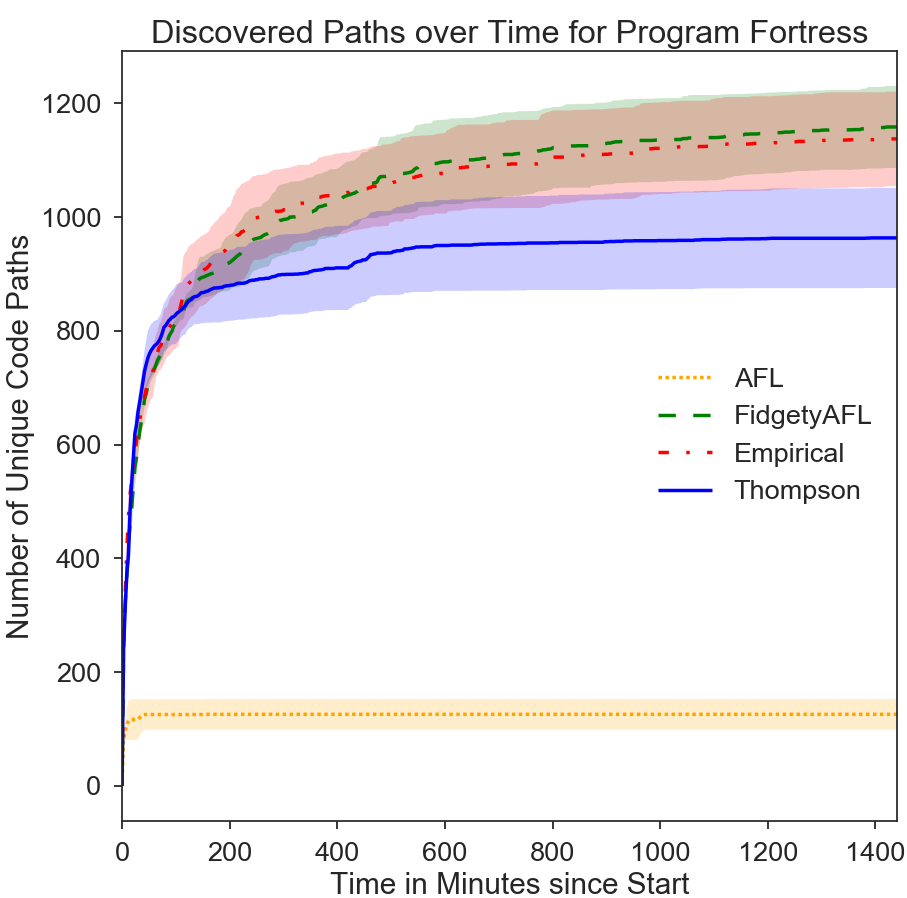}
\caption{Example 24hr runs on test programs (plotted with mean/stdev across 3 trials). At the beginning, the fuzzers discover many new code paths, but as time goes on productivity decreases, until a new code path unlocks further discovery. On the left we have a typical run where Thompson sampling is most productive, on program \texttt{ASCII_Content_Server} from Section \ref{sec:motivation}. In the middle, we have a run where Thompson sampling is extremely effective at the beginning of fuzzing, with the Empirical distribution fuzzer gaining an edge later on. On the right, we have a run where FidgetyAFL is most effective.}
\label{fig:graphs}
\end{figure*}

For our Thompson Sampling experiments, we need to resolve the Credit Assignment problem from Section \ref{sec:credit}, by finding a fixed (and ideally small) return value for \texttt{sample\_num\_mutations} (line 8 of Algorithm \ref{alg:greybox}). To estimate this value, we propose the following experiment: Let $n$ be the number of stacked mutations that AFL runs. Run multiple trials of AFL for a fixed amount of time, with fixed values of $n$, in the range of values that AFL usually samples from (recall that AFL samples uniformly among $\{2^0, 2^1, \ldots 2^7\}$). Then, pick the $n$ with the performance closest to (or better than) standard AFL.

The results in Table \ref{tab:nstack} show relative coverage statistics across different fixed values of $n$ every 30 minutes in a 2-hr fuzzing run. Note that we fix values of $n$ for FidgetyAFL instead of regular AFL, as like prior work, we found FidgetyAFL to consistently perform much better that AFL. These statistics were collected on the set of 75 training CGC binaries. It's extremely interesting to note that fixed values of $n = 2$, $n = 4$, and $n = 8$ each show better performance than FidgetyAFL, which samples uniformly from a range of values. Looking further at the data, we decided to use $n=4$ to build our approach, as it performs slightly better than $n = 2$ and $n = 8$. 

\subsection{Learning a Base Empirical Distribution}
\label{sec:empres}

Next, we estimate a good stationary distribution over mutation operators, as per Section \ref{sec:empirical}. We estimate this ``Empirical'' distribution (as we shall refer to it through the rest of this paper) utilizing the procedure outlined in Section \ref{sec:empirical}; namely, we run FidgetyAFL on our training binaries, and we count how many times each mutation operator was involved in the creation of a successful input (improving code coverage). We then normalize these counts to come up with our final Empirical distribution. We perform our training runs by fuzzing the 75 training binaries with FidgetyAFL for a 24 hour period, dumping logging statistics every 3 minutes. Results with the Empirical distribution can be found in the next section.

\subsection{24 Hour Experiments}
\label{sec:results}

We run AFL, FidgetyAFL, the Empirical distribution fuzzer, and the Thompson Sampling fuzzer on the DARPA CGC and the LAVA-M binaries for 24 hours each. The Thompson Sampling experiments follow the procedure outlined in Section \ref{sec:thompson}. Utilizing the fixed value of $n = 4$ from Section \ref{sec:num-stacks}, we estimate the parameters $\theta_k$ from Section \ref{sec:thompson} directly from data collected during the fuzzing run. We start each Thompson Sampling run with a uniform mutator distribution, as well as a uniform prior (we utilize a Beta distribution with $\alpha = 1$ and $\beta = 1000$ as the prior for each arm). Then, every 10 minutes, we update the distribution by sampling from the updated posterior, estimated using the counts of successful/unsuccessful operators we have observed thus far. Results are as follows:

\subsubsection{DARPA Cyber Grand Challenge}

Table \ref{tbl:darpa-res} contains the results for the DARPA CGC experiments. We report relative coverage numbers at the 6, 12, 18, and 24 hour marks, as well as aggregate win counts. Furthermore, we report the number of unique crashes found at the end of the 24 hour fuzzing period. Recall that unique crashes are inputs that exercise a unique code path that result in a crash -- for each bug in a program, there can be possibly hundreds of unique crashes (as you can reach the same bug by traversing different program paths). We present graphs showing progress over time for three different CGC binaries in Figure \ref{fig:graphs}. 

\begin{table}[b]
\begin{tabular}{@{}lcccc@{}}
\toprule
               & \texttt{base64}     & \texttt{md5sum}     & \texttt{uniq}     & \texttt{who}        \\ \midrule
AFL            & 117 $\pm$ 20 & 55 $\pm$ 2   & 13 $\pm$ 0 & 37 $\pm$ 1   \\
FidgetyAFL     & 133 $\pm$ 10 & 340 $\pm$ 10 & 87 $\pm$ 1 & 372 $\pm$ 36 \\
Empirical & 134 $\pm$ 8  & 406 $\pm$ 22 & 80 $\pm$ 1 & 115 $\pm$ 9  \\
Thompson       & 144 $\pm$ 14 & 405 $\pm$ 2  & 75 $\pm$ 2 & 106 $\pm$ 16 \\ 
FairFuzz       &  138 $\pm$  7  &  348 $\pm$ 5  & 70 $\pm$ 3         & 232 $\pm$ 18            \\
FairFuzz + Thompson      &  85 $\pm$ 6   &   127 $\pm$ 21   &   50 $\pm$ 4       &  23 $\pm$ 11           \\\bottomrule
\end{tabular}
\caption{Number of code paths discovered for each LAVA-M binary, after 24 hours. Numbers reported are mean/standard error across three separate trials.}
\label{tbl:lava-paths}
\end{table}

\subsubsection{LAVA-M}

Table \ref{tbl:lava-paths} reports the mean and standard error across number of code paths discovered for each strategy, across 3 separate 24-hour trials (we ran extra trials here because of the small size of the dataset). Table \ref{tbl:lava-crashes} reports the mean/standard error across the number of verifiable bugs discovered by each fuzzer, again across 3 separate 24-hour trials (here, LAVA-M gives us the ability to de-dup crashes, so we report bugs instead of crashes). 

\begin{table}[b]
\begin{tabular}{@{}lcccc@{}}
\toprule
               & \texttt{base64}     & \texttt{md5sum}     & \texttt{uniq}     & \texttt{who}        \\ \midrule
AFL            & 15 $\pm$ 5  & 0 $\pm$ 0   & 0 $\pm$ 0 & 0 $\pm$ 0   \\
FidgetyAFL     & 26 $\pm$ 9  & 4 $\pm$ 1   & 1 $\pm$ 1 & 201 $\pm$ 56 \\
Empirical      & 22 $\pm$ 5  & 0 $\pm$ 0   & 0 $\pm$ 0 & 78 $\pm$ 17  \\
Thompson       & 31 $\pm$ 8  & 1 $\pm$ 1   & 0 $\pm$ 0 & 106 $\pm$ 16 \\ 
FairFuzz       &   18 $\pm$ 4         &  1 $\pm$  1          &  0 $\pm$  0        &  18 $\pm$ 5            \\
FairFuzz + Thompson      &   5 $\pm$ 2         &  0 $\pm$ 0           &  0 $\pm$ 0          &   0 $\pm$  0          \\\bottomrule
\end{tabular}
\caption{Number of unique bugs (de-duped) discovered for each LAVA-M binary, after 24 hours. Numbers reported are mean/standard error across three separate trials.}
\label{tbl:lava-crashes}
\end{table}

\subsection{Thompson Sampling and FairFuzz}

FairFuzz \cite{Lemieux2017FairFuzzTR} is a recently introduced tool built on top of AFL, optimizing AFL's core functionality with a smarter, data-driven approach for learning the \texttt{pick_input} and \texttt{sample_mutation_site} functions from Algorithm \ref{alg:greybox}. Specifically, FairFuzz improves fuzzing efficiency by prioritizing queue inputs that exercise \textit{rare branches}, or parts of the program under test that aren't touched by many of the discovered inputs. Furthermore, FairFuzz performs an analysis of which bytes in the input trigger certain branch behavior -- a sort of estimated taint tracking. In doing so, FairFuzz learns a mask over input bytes that forces AFL to leave certain bytes alone (i.e. headers or checksums), while focusing mutations on more volatile sections. In the original FairFuzz paper \cite{Lemieux2017FairFuzzTR}, the authors show that their approach is significantly more efficient than AFL, FidgetyAFL, and another AFL-variant, AFLFast \cite{Bhme2016CoveragebasedGF}. 

With this in mind, we wanted to answer two additional questions: 1) does Thompson Sampling approach provide better benefits than those behind FairFuzz, and 2) are the optimizations behind FairFuzz are complementary to Thompson Sampling. To explore this, we performed additional experiments, running FairFuzz (with the recommended configurations) on our test set of 75 DARPA CGC binaries. Furthermore, we ported our Thompson Sampling changes to FairFuzz, and ran this new FairFuzz + Thompson Sampling fuzzer on the same binaries. Finally, we compared the three strategies in terms of relative coverage, and unique crashes discovered. Table \ref{tbl:fairfuzz} reports the results for the FairFuzz experiments, after fuzzing each of the 75 CGC test binaries for 24 hours. Thompson sampling seems to have a significant edge in both crash finding and code coverage, finding around 500 more crashes than FairFuzz across all binaries, and obtaining 7\% more coverage. What is especially interesting, though, is that there is a significant dip in performance when combining the FairFuzz optimizations with the Thompson Sampling optimizations over estimating a mutation operator distribution. Tables \ref{tbl:lava-paths} and \ref{tbl:lava-crashes} show FairFuzz results on the LAVA-M binaries; the results are again muddled, with Thompson Sampling outperforming FairFuzz on 3 of the 4 binaries. However, these experiments echo the CGC experiments, in that there is a significant dip in performance for the combined FairFuzz + Thompson fuzzer.

\begin{table}[b]
\begin{tabular}{@{}lccc@{}}
\toprule
                    & 24 hr         & Crashes & Wins / All \\ \midrule
FairFuzz            & 0.88 $\pm$ 0.02 & 734     & 17         \\
\textbf{Thompson}            & \textbf{0.95 $\pm$ 0.01} & \textbf{1287}    & \textbf{49}         \\
\textit{FairFuzz + Thompson} &\textit{ 0.57 $\pm$ 0.03} & \textit{245}     & \textit{1}          \\ \bottomrule
\end{tabular}
\caption{Results for the FairFuzz experiments, on the 75 test CGC binaries. Wins are reported excluding ties.}
\label{tbl:fairfuzz}
\end{table}

\begin{figure}
\includegraphics[width=.7\linewidth]{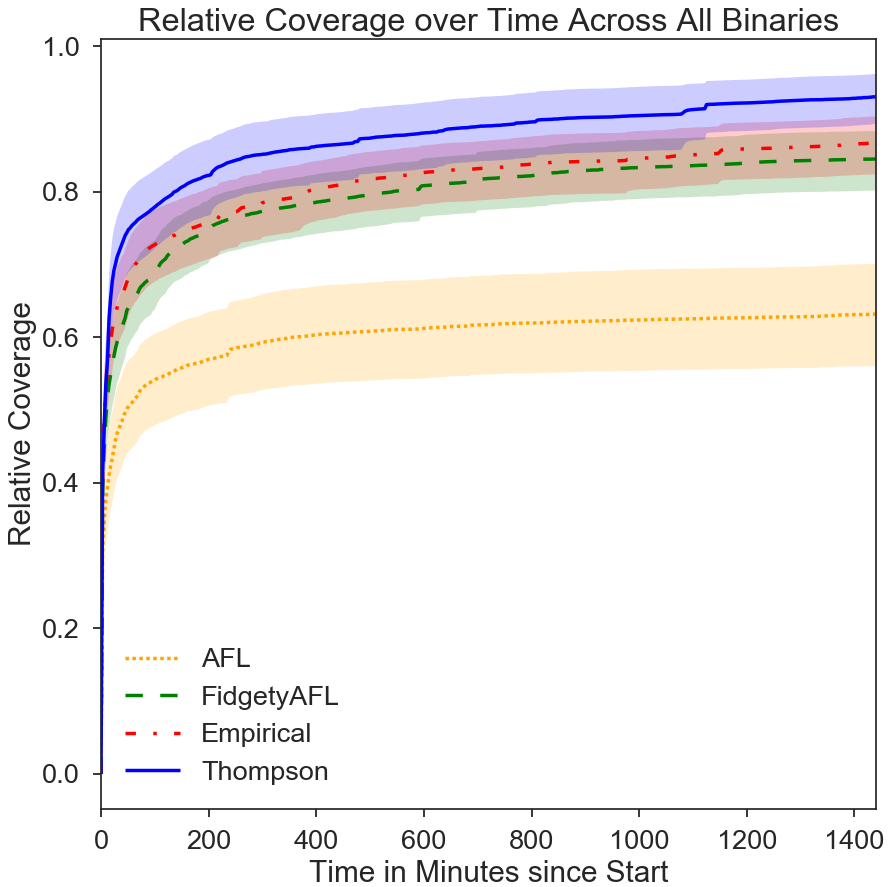}
\caption{Summary graph, showing relative code coverage over time, across all programs (CGC + LAVA-M). Plot above is depicted with error bars, representing a 95\% confidence interval (across programs).}
\label{fig:summary}
\end{figure}

\section{Discussion}
\label{sec:discussion}

Overall, the Thompson Sampling approach performed very well, beating the best baselines, FidgetyAFL and FairFuzz by a large margin across the vast majority of binaries. Especially on the DARPA CGC binaries, the Thompson Sampling fuzzer shows an almost 10 percent improvement in relative coverage over FidgetyAFL, while also discovering 1336 unique crashes across all 75 test binaries, almost double the amount found by any other fuzzer. Additionally, of the 75 binaries, it finds more code paths than FidgetyAFL on 52 binaries (excluding those binaries on which both strategies discovered the same number of paths). Looking at the graphs in Figure \ref{fig:graphs}, we see that Thompson Sampling in the usual case beats FidgetyAFL/AFL by a relatively significant margin (in the case of programs like \texttt{ASCII_Content_Server}, and in some cases, can find thousands more code paths (in the case of programs like \texttt{CML}). In addition to the Thompson Sampling results, we also see that the estimated Empirical distribution fuzzer performed rather well, beating over FidgetyAFL by around 3 percent in terms of relative coverage. That being said, on the DARPA CGC binaries, it is unable to find more crashes than the baselines. Figure \ref{fig:summary} shows a summary graph of relative coverage over time, evaluating Thompson Sampling, the Empirical Distribution, and both versions of AFL across all binaries (CGC + LAVA-M), with error bars representing a 95\% confidence interval across programs. We see that Thompson Sampling outperforms the other strategies at every timestep, and as fuzzing continues, it outperforms other strategies by a larger margin.

Thompson Sampling is not a silver bullet, however; there are some programs on which it fails to overcome FidgetyAFL (namely, on the 14 programs mentioned in Table \ref{tbl:darpa-res}, and on the last graph in Figure \ref{fig:graphs}). This limitation is further explored with the LAVA-M binaries, where we see that the Thompson Sampling approach obtains lukewarm results. While on binaries \texttt{base64} and \texttt{md5sum}, the Thompson sampling approach is able to find more code paths, on the remaining two binaries \texttt{uniq} and \texttt{who}, FidgetyAFL wins, in some cases by a large margin. A possible explanation for this is due to the credit assignment adjustment we made for Thompson Sampling -- by fixing the number of mutation operators to a small constant (in our case, $4$ mutations run per input), we are in fact limiting the expressive power of our fuzzer. The FidgetyAFL baseline, however, is unconstrained, and in some cases can create inputs that differ from their parents by as many as 128 mutations, thereby triggering certain parts of the code that the limited mutations cannot. This opens up a possible avenue for future work; getting around the credit assignment problem and learning adaptive distributions across \textit{any} choice of \texttt{sample_num_mutations} would allow for fuzzers more powerful than those developed in this work. Additionally, when it comes to crashes, we see a similar trend, with Thompson sampling only finding more on the \texttt{base64} binary. However, it is important to note the high variance across crashes -- this is again indicative of the synthetic nature of the LAVA-M dataset. Because each bug is triggered only by a single setting of 4 bytes of the input buffer, it is quite random whether a given run finds certain bugs or not. This seems to suggest that the LAVA-M binaries are more conducive to testing symbolic execution tools and white box fuzzers that actually inspect branch conditions in the code, and symbolically solve for inputs to pass certain conditions; for random processes like AFL and AFL-variants, it does not provide the most interpretable and conclusive results.

Finally, compared to FairFuzz, we see that Thompson Sampling has a significant advantage, in terms of relative code coverage and crashes found across all 75 CGC binaries. This again indicates the role the mutation distribution has on the efficacy of grey-box fuzzing, compared to other optimizations like picking queue inputs. What is especially interesting is that when combining the two seemingly orthogonal optimizations (FairFuzz's input and mutation site selection, with Thompson Sampling's mutation operator selection), we see a significant drop in performance. While unfortunate, a possible explanation is due to the disconnect between how FairFuzz picks inputs, and how Thompson Sampling estimates its distribution over mutators. FairFuzz prioritizes inputs that exercise rare branches of the program -- inputs that make up only a slight percentage of the total queue. However, Thompson Sampling estimates its distribution over mutators based on \textit{all} inputs in the queue. As a result, the Thompson Sampling distribution is not the same as the \textit{optimal} distribution for the FairFuzz chosen inputs. To remedy this, a possible avenue of future work is to learn an input-conditional distribution over mutators -- by learning a different distribution for each input, one can learn a more fine-grained policy for choosing mutators, circumventing this distributional asymmetry.

\section{Conclusion}


In this work, we have presented a system for combining machine learning and bandit-based optimization with state-of-the-art techniques in grey-box fuzzing. Utilizing Thompson Sampling to adaptively learn distributions over mutation operators, we are able to show significant improvements over many state-of-the-art grey-box fuzzers, like AFL, FidgetyAFL, and the recently released FairFuzz. While our approach is not a silver bullet that gains optimal results on all programs fuzzed, it works in the vast majority of cases across many real-world, complex programs, and in some cases, by several orders of magnitude, finding hundreds more code paths and crashes than other baselines. Our experiments conclusively show that there are significant gains to be made by refining existing fuzzing tools (and developing new ones!) with data-driven machine learning techniques for adapting to wide varieties of different programs.

\bibliographystyle{ACM-Reference-Format}
\bibliography{references}

\end{document}